\documentclass[letterpaper]{article} 
\usepackage{aaai25}  
\usepackage{times}  
\usepackage{helvet}  
\usepackage{courier}  
\usepackage[hyphens]{url}  
\usepackage{graphicx} 
\usepackage{natbib}  
\usepackage{caption} 
\usepackage[ruled]{algorithm2e}
\SetKwComment{Comment}{// }{}
\usepackage{graphicx}
\usepackage{booktabs}
\usepackage{custom_style}
\usepackage{float}
\usepackage{multirow}
\usepackage{adjustbox}
\usepackage{subcaption}
\usepackage{thm-restate}  
\usepackage{cleveref}
\usepackage{newfloat}
\usepackage{listings}
\DeclareCaptionStyle{ruled}{labelfont=normalfont,labelsep=colon,strut=off} 
\floatstyle{ruled}
\newfloat{listing}{tb}{lst}{}
\floatname{listing}{Listing}
\newcommand{\minisection}[1]{\noindent{\textbf{#1}}}

\title{Self-Corrected Flow Distillation for Consistent One-Step and Few-Step Text-to-Image Generation}
\author {
    Quan Dao\equalcontrib \textsuperscript{\rm \rm 1,\rm 2\rm \textdagger}, 
    Hao Phung\equalcontrib \textsuperscript{\rm 1,\rm 3\rm \textdagger}, 
    Trung Tuan Dao\textsuperscript{\rm 1}, 
    Dimitris N. Metaxas\textsuperscript{\rm 2}, 
    Anh Tran\textsuperscript{\rm 1}
}
\affiliations {
    \textsuperscript{\rm 1}VinAI Research\\ \textsuperscript{\rm 2}Rutgers University\\ \textsuperscript{\rm 3}Cornell University \\
    quan.dao@rutgers.edu,  htp26@cornell.edu,  v.trungdt21@vinai.io,  dnm@cs.rutgers.edu,  v.anhtt152@vinai.io
}

\usepackage{bibentry}

\begin{document}

\maketitle
\def\thefootnote{\textsuperscript{\textdagger}}\footnotetext{Work done while at VinAI.}

\begin{abstract}
Flow matching has emerged as a promising framework for training generative models, demonstrating impressive empirical performance while offering relative ease of training compared to diffusion-based models. However, this method still requires numerous function evaluations in the sampling process. To address these limitations, we introduce a self-corrected flow distillation method that effectively integrates consistency models and adversarial training within the flow-matching framework. This work is a pioneer in achieving consistent generation quality in both few-step and one-step sampling. Our extensive experiments validate the effectiveness of our method, yielding superior results both quantitatively and qualitatively on CelebA-HQ and zero-shot benchmarks on the COCO dataset. 

\end{abstract}

\begin{links}
\link{Code}{https://github.com/hao-pt/SCFlow.git}
\end{links}

\section{Introduction}
The field of generative modeling has witnessed remarkable progress over the past decade. The modern generative models could create diverse and realistic content across various modalities. Previously, Generative Adversarial Networks (GANs) \cite{goodfellow2014generative, karras2019style} was dominant in this field by their ability to create realistic images. However, training GAN models is costly in both time and resources due to training instability and mode collapse. The emergence of diffusion models \cite{ho2020denoising, song2019generative, song2020score} marked a significant focus shift in generative AI. These models, exemplified by groundbreaking works such as DALL-E \cite{ramesh2021zero} and Stable Diffusion \cite{rombach2022high} have surpassed GANs to become the current state-of-the-art in image synthesis. Diffusion models define a fix forward process which gradually perturbs image to noise and learn a model to perform the reverse process from noise to image. Their success lies in the ability to capture complex distribution of data and produce fidelity and diverse images. This approach has effectively addressed many of the limitations faced by GANs, offering improved stability, diversity, and scalability. However, diffusion training takes long time to converge and requires many NFEs to produce high-quality samples. Recent works \cite{lipman2023flow, liu2022rectified, albergo2022building} have introduced a flow matching framework, which is motivated by the continuous normalizing flow. By learning probability flow between noise and data distributions, flow matching models provide a novel perspective on generative modeling. Recent advancements have demonstrated that flow matching can achieve competitive results with diffusion models \cite{ma2024sit} while potentially offering faster sampling.

\begin{figure}[t]
    \centering
    \includegraphics[width=0.9\linewidth]{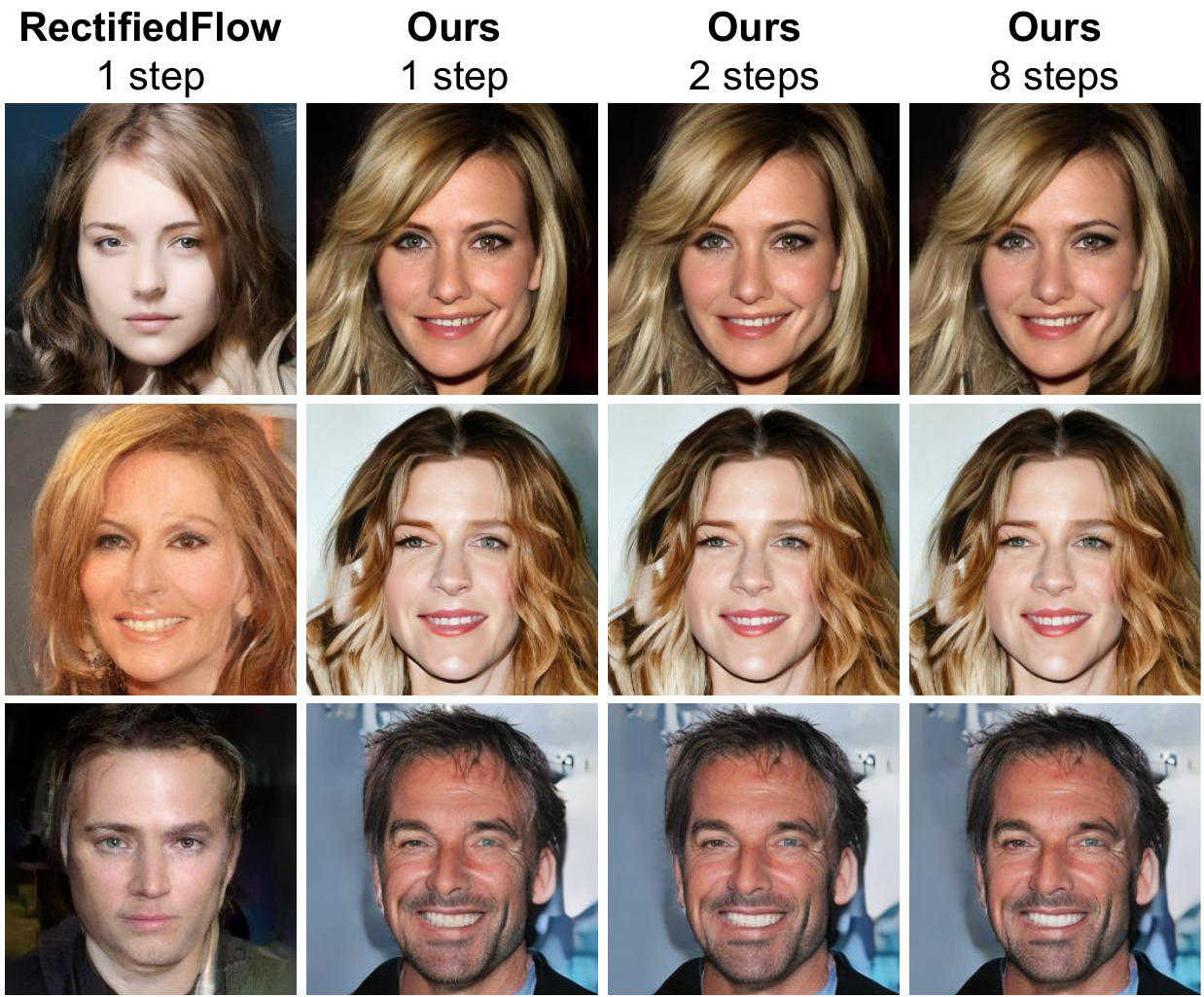}
    \caption{Illustration of consistent one-step and few-step image generation. Our method consistently delivers superior visual quality across different sampling steps, significantly surpassing the performance of the RectifiedFlow counterpart.}
    \label{fig:teaser}
\end{figure}

While flow matching can generate high-quality images with fewer NFEs compared to diffusion models, it still shares the challenge of prolonged sampling times due to its inherently iterative denoising process. This limitation poses a significant barrier to the practical application of both flow matching and diffusion models in real-world scenarios. To address this challenge in diffusion models, recent works \cite{meng2023distillation, luo2023latent, gu2023boot, nguyen2024swiftbrush, dao2024swiftbrush, sauer2023adversarial, xu2024ufogen} focus on developing timestep distillation technique and show remarkable results. For example, LCM \cite{luo2023latent} utilizes consistency distillation \cite{song2023consistency} and yields good results but generates blurry images at one-step sampling. SwiftBrush \cite{nguyen2024swiftbrush} adopts SDS loss for distillation into a one-step generator but sacrifices the ability to perform multi-step sampling. Both UFOGEN \cite{xu2024ufogen} and SD Turbo \cite{sauer2023adversarial} are able to generate high-quality images with one and few-step sampling. However, these methods struggle to maintain consistent results across different sampling schemes. In the context of flow matching, InstaFlow \cite{liu2023instaflow, liu2022rectified} addresses this issue by utilizing rectified flow to produce direct transitions from source to target data. Instaflow has three training stages: collecting data, rectified flow and distillation. The instaflow could produce high-quality one-step generation but fail to perform few-step sampling due to their simple regression distillation at the third stage. 

\renewcommand{\thefootnote}{\arabic{footnote}}

In this paper, we investigate how to distill a latent flow matching teacher into a consistent one and few-step generator. Motivated from consistency model \cite{luo2021diffusion, song2023consistency}, we apply consistency framework into latent flow model. However, we found that naively applying consistency distillation faces two challenges which are blurry one-step and oversaturated few-step generated images\footnote{ ``oversaturated'' refers to the phenomenon where images generated by the model exhibit excessively vibrant colors and overly high contrast, resulting in a loss of natural color balance and detail.}. The blurry one-step is also observed in LCM \cite{luo2021diffusion}. These limitations could be due to discrepancy in statistic of latent compared to pixel space. To deal with blurry one-step generation, we propose to use GAN model for enhancing quality of one-step images. For oversaturated few-step problem, we introduce truncated consistency and reflow loss. These losses effectively mitigate the oversaturation problem, ensuring improved performance in few-step sampling. Besides, we also propose bidirection loss to improve the consistency across different sampling schemes. Our proposed framework is called self-corrected flow distillation. By thorough experiments, we validate the effectiveness of our framework to produce high quality and consistent images in both one and few-step sampling.

Our key contributions are threefold:
\begin{itemize}
    \item We propose a training framework to effectively address the unique challenges of latent consistency distillation and offers optimal combinations for improved performance, including a truncated consistency loss to mitigate oversaturation, GAN to overcome blurry one-step generation. Additionally, the reflow and bidirection losses are introduced to enhance the consistency of generator across different sampling steps. 
    \item Through extensive experiments on multiple datasets, we demonstrate that our approach significantly outperforms existing methods in both one-step and few-step generation, achieving competitive FID scores while maintaining generation speed. We provide detailed ablation studies to analyze the impact of each component of our method.
    \item For the first time, we have achieved consistent, high-quality image generation in both few-step and one-step sampling using flow matching. The model will be publicly released to support further research.
\end{itemize}

\section{Related Work}

\subsection{Flow Matching}
Flow matching is emerging as the competitive alternative to diffusion models, as it deterministically finds the mapping between noise and data distribution. The deterministic property is favored in many generative applications, such as image inversion \cite{pokle2023training} and editing \cite{hu2024lfm}, as well as in video and beyond \cite{davtyan2023efficient, song2024equivariant, gao2024lumin-t2x}, due to its fast generation capability and reduced need for large NFE \cite{liu2023flow, lipman2023flow, liu2022rectified, dao2023flow}. Recently, some works has linked the connection between diffusion models (known as score-based models) and flow matching \cite{kingma2024understanding, ma2024sit}. Given these advantages, SDv3 \cite{esser2024scaling} has adopted flow matching as their core framework, combined with a powerful transformer-based architecture \cite{Peebles2022DiT}, resulting in groundbreaking image generation capabilities. However, the computational complexity of iterative sampling still hinders these models from achieving real-time performance and lags behind GAN counterparts \cite{kang2023gigagan, sauer2023stylegan}. Therefore, developing one-step and few-step sampling techniques is crucial to strike a balance between generation quality and sampling speed.

\begin{figure}[t]
    \centering
    \includegraphics[width=\linewidth]{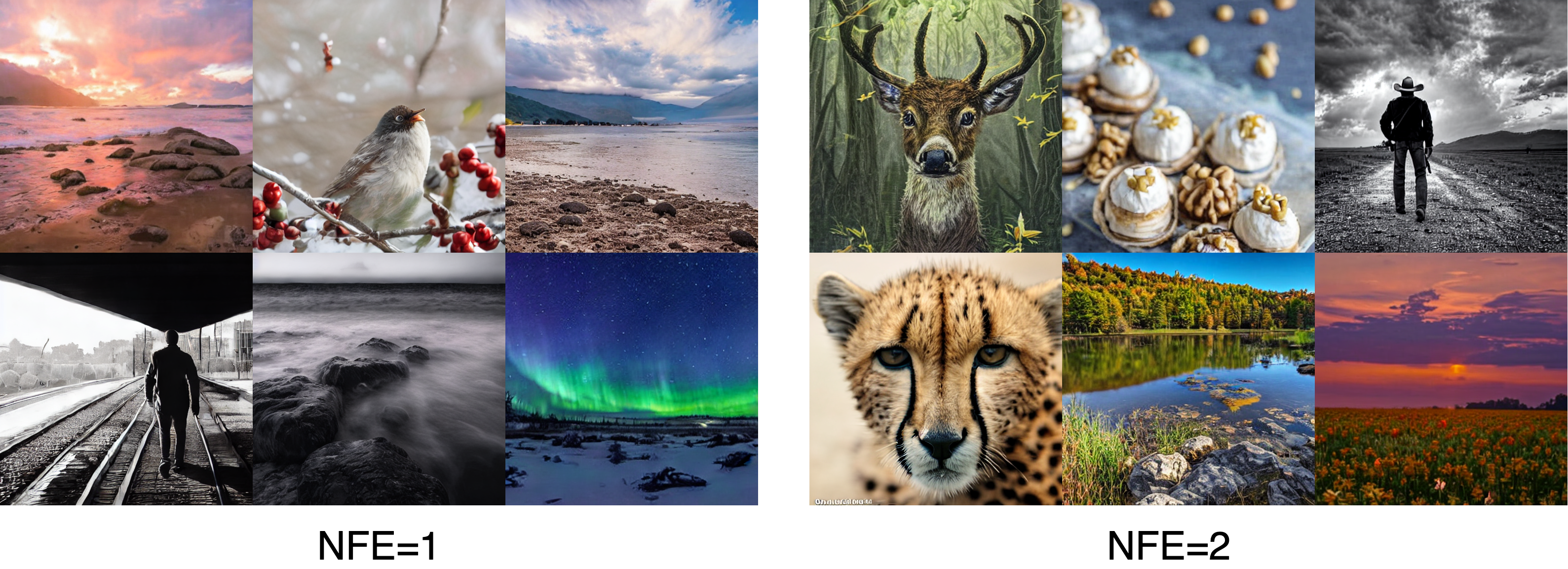}
    \caption{Qualitative results of our Distilled Text-to-Image diffusion model.}
    \label{fig:t2i_teaser}
\end{figure}

\begin{figure*}[t]
    \centering
    \includegraphics[width=0.7\linewidth]{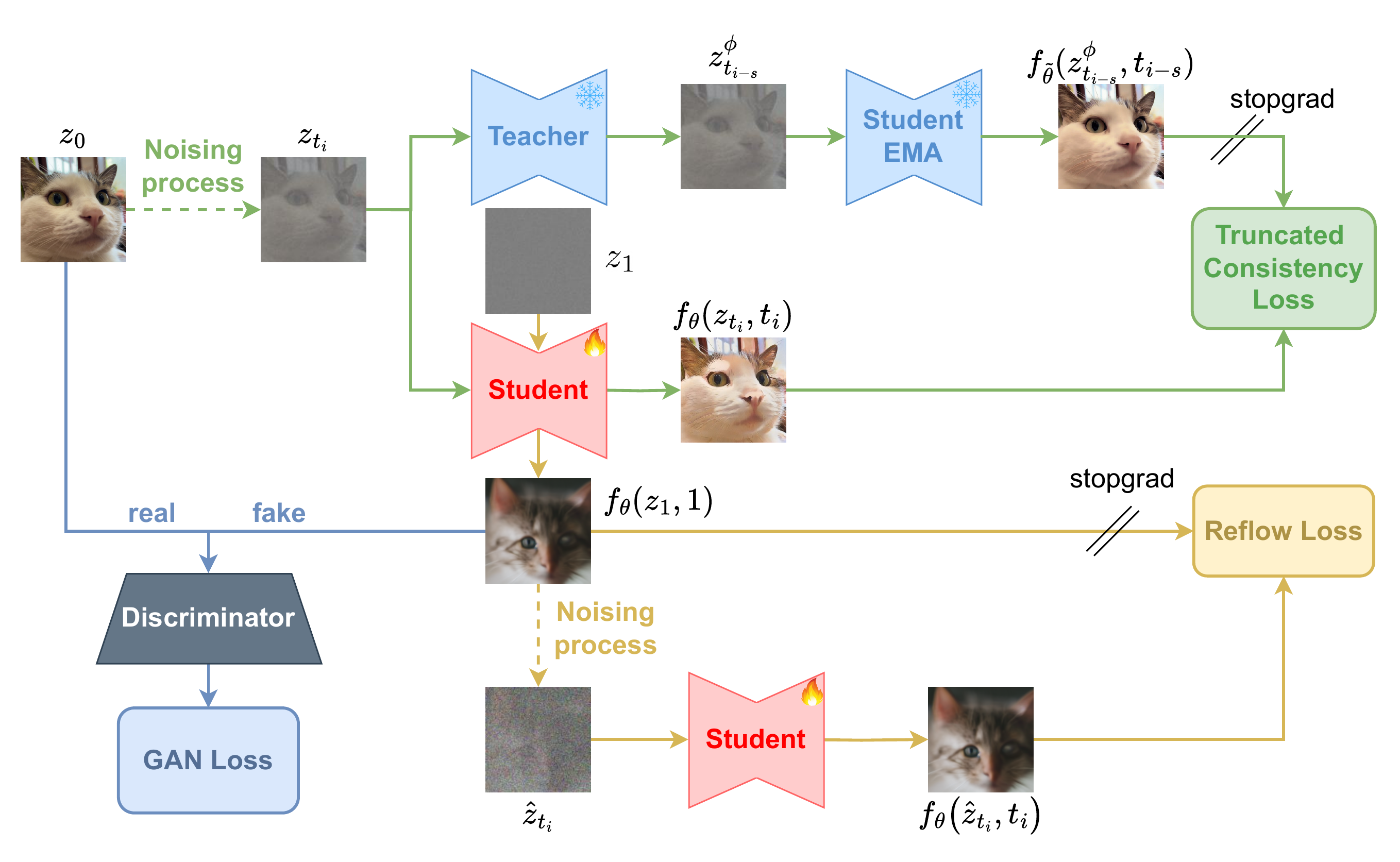}
    \caption{Overview of the proposed Self-Corrected Flow Distillation framework. For clarity, all latent representations are visualized as images. Our method builds upon latent consistency distillation and introduces three key components. First, a truncated consistency loss mitigates the oversaturation artifacts that arise at small timesteps (where $t \to 0$ and the noisy latent approaches the clean data), at which the consistency loss produces near-zero gradients that offer insufficient learning signal. Second, an adversarial objective (GAN loss) encourages one-step generations to remain faithful to real images, thereby alleviating the blurriness commonly observed in single-step inference. Third, since the above losses alone do not fully resolve oversaturation — particularly the quality gap between one-step and few-step generation — we propose a reflow loss that enables the model to self-correct its flow estimation, anchored to the reliable one-step prediction $f_\theta(z_1, 1)$ whose quality is reinforced by the GAN objective. Additionally, we introduce bi-directional consistency distillation to enforce consistency at both endpoints of the denoising trajectory.}
    \label{fig:systemfig}
\end{figure*}

\subsection{Distillation Technique}
Knowledge distillation \cite{hinton2015distilling} has gained remarkable success in enhancing the performance of lightweight models under the guidance of a complex teacher model, allowing the student model to match or even surpass the teacher one. In context of diffusion models, instead of reducing model size, there is a line of methods \cite{luhman2021knowledge,salimans2022progressive} that aims to distill a pre-trained diffusion-model teacher for reducing the number of sampling steps. Recently, Consistency model \cite{song2023consistency} has achieved promising results in enhancing sampling efficiency of diffusion models. LCM \cite{luo2023latent} is the closest to our work where they directly adopted the consistency distillation objective, allowing Stable Diffusion model \footnote{https://github.com/Stability-AI/stablediffusion.git} to generate an image with just a few steps. Similar to our work, their method is also exploited in latent space of a pre-trained encoder. In contrast, our method proposes a consistency-based distillation method that is well adapted to the flow-matching framework. By combining with adversarial training and reflow objectives, our method can significantly increase the performance of few-step generation for both unconditional and conditional tasks. Unlike RectifiedFlow \cite{liu2022rectified}, which iteratively fine-tunes the model on generated noise-image pairs using a pretrained flow model, our method eliminates the need for this costly, separate flow-straightening stage for distillation. Instead, it performs flow rectification and distillation concurrently during only one-stage training.
While our method integrates adversarial objectives akin to AdversarialDSM \cite{jolicoeur2020adversarial}, it distinguishes itself by optimizing GAN on latent encoded features instead of coarse pixels. 
This enhances the generation quality of one-step generation, diverging from a sole focus on improving the fidelity of score-based networks like the former.

\section{Method}
Flow matching exhibits faster training convergence \cite{dao2023flow} and better image generation \cite{ma2024sit} compared to diffusion model. Thanks to these advantage, research community start shifting attention to this framework. Recent work \cite{esser2024scaling} has scaled up the flow matching to text-to-image generation with high quality result. In contrast, flow matching still take long time for sampling compared to GAN model. This motivates us take deeper investigation in distillation method for this framework, as illustrated in \Cref{fig:systemfig}.

In this section, we start by revisiting latent flow matching framework \cite{ma2024sit, dao2023flow}.  Next, we detail technical aspects of our proposed distillation framework named Self-Corrected Flow Distillation. Our distillation method is motivated from \cite{luo2023latent, song2023consistency}. We show that straightly applying consistency distillation on latent flow matching framework yield low quality generation with both one-step and few-step sampling scheme. This behaviour also appeared in LCM \cite{luo2023latent}, which remains unsolved until now. By utilizing GAN and Rectified technique \cite{liu2022rectified}, we could mitigate the drawbacks of latent consistency distillation.

\subsection{Preliminary}

\begin{algorithm}[t]
\caption{Self-Corrected Flow Distillation}\label{alg:utrain}
\KwData{data $p_0$, Encoder $\cE$, distilled model $v_\theta$, pretrained model $v_\phi$, lr $\eta$, ema decay $\mu$, and $\lambda_{GAN},\lambda_{RF},\lambda_{BI}$ are weight terms} 

$\theta \gets \phi$ \;

\For{$iter \in \{1,\dots,N\}$}{
    $\bx_0 \sim p_0$, $\bz_0 \gets \mathcal{E}(\bx_0)$\; 
    $\bz_1 \gets \cN(0, \bI)$, $i \sim \cU[1, N]$\;
    $\bz_{t_i} \gets (1 - t_i)\bz_0 + t_i\bz_1$\;

    $\mathcal{L}_{distill} = \mathcal{L}_{CD} $ (using \cref{eq:L_CD})
    
    \If {$iter \geq N_{GAN}$}{
         $\mathcal{L}_{distill} \gets \mathcal{L}_{distill} + \lambda_{GAN}*\mathcal{L}_{GAN} $ (using \cref{eq:L_GAN})}

    \If {$iter \geq N_{RF}$}{
         $\mathcal{L}_{distill} \gets \mathcal{L}_{distill} + \lambda_{RF}*\mathcal{L}_{RF} $ (using \cref{eq:L_RF})}

    \If {$iter \geq N_{BI}$}{
         $\mathcal{L}_{distill} \gets \mathcal{L}_{distill} + \lambda_{BI}*\mathcal{L}_{BI} $ (using \cref{eq:L_Bi})}
    
    $\theta \gets \theta-\eta \nabla_{\theta} \mathcal{L}_{distill}$\;
    $\tilde\theta = \mathbf{sg}\left( \mu \tilde\theta + (1 - \mu)\theta \right)$
}\label{alg:distill}
\end{algorithm}
 
Given the training dataset $\mathbf{D}$, we draw a sample $\mathbf{x}_0 \in \mathbf{R}^{d}$ from the dataset. Denote that $\mathcal{E}$ and $\mathcal{D}$ are encoder and decoder of a pretrained VAE, we obtain the latent $\mathbf{z}_0 = \mathcal{E}(\mathbf{x}_0) \in R^{d/h}$, where $h$ represents the compressed rate of VAE model. The training objective of latent flow matching is to approximate a probabilistic path from a random noise $\mathbf{z}_1 \sim \mathcal{N}(0, \mathbf{I}^{d/h})$ to the training dataset distribution $\mathbf{z}_0$. Previous works \cite{ma2024sit,dao2023flow,liu2022rectified,lipman2023flow} use the following velocity loss to train flow matching framework:
\begin{equation}
  \hat{\theta} = \argmin_{\theta} \mathbf{E}_{t, \bz_t} \left[\norm{\bz_1 - \bz_0 - v_\theta\left(\bz_t, t\right) }^2_2 \right].
  \label{eq:fm-obj-constant}  
\end{equation}

To enable the conditional generation, the condition information $\mathbf{c}$ is injected into the flow matching framework as below \cite{liu2023instaflow,ma2024sit,dao2023flow}:
\begin{equation}
  \hat{\theta} = \argmin_{\theta} \mathbf{E}_{t, \bz_t} \left[\norm{\bz_1 - \bz_0 - v_\theta\left(\bz_t, \bc, t\right) }^2_2 \right].
  \label{eq:clfm-obj-constant}
\end{equation}

Conditioning information $c$ can be images, text, or class labels, with different conditional mechanisms like AdaIN \cite{Peebles2022DiT} or cross-attention \cite{wang2018high}.

To better control the diversity and quality of generation, previous works \cite{liu2023instaflow, dao2023flow} adopt classifier-free guidance sampling algorithm similar to \cite{ho2022classifier}:
\begin{equation}
\tilde{v}_{\theta}(\bx_t, \bc, t) \approx \gamma v_{\theta}(\bx_t, \bc, t) + (1 - \gamma) v_{\theta}(\bx_t, \bc=\emptyset, t),
\label{eq:vel_free_sampling}
\end{equation}
where $v_{\theta}(\bx_t, \bc=\emptyset, t)$ represents the unconditional velocity trained with null token $\bc$. Hyperparameter $\gamma$ controls the generation of flow matching framework. While smaller values of $\gamma$ promote diverse outputs, larger $\gamma$ values tend to yield higher fidelity images at the cost of reduced diversity.

\subsection{Self-Corrected Flow Distillation}

Given pretrained latent flow matching model $v_\phi$, we would like to distill from that teacher model to a student $v_\theta$ that is capable of both one or many step sampling. Therefore, we firstly apply consistency distillation \cite{luo2023latent, song2023consistency} for pretrained teacher over $N$ discrete times $0=t_1 < t_2 < \dots < t_N = 1$ as follows:
\begin{equation}
    \mathcal{L}_{CD} = \mathbf{E}_{t_i, \bz_{t_i}} \left[\norm{f_{\tilde\theta} \left( \bz^{\phi}_{t_{i-s}}, t_{i-s} \right) - f_\theta\left(\bz_{t_i}, t_i\right) }^2_2 \right],
\end{equation}
where $s$ is skipping timesteps and $\tilde\theta = \mathbf{sg}\left( \mu \tilde\theta + (1 - \mu)\theta \right)$ is the exponential moving average (EMA) of $\theta$ model with a decay rate $\mu \in \left[0, 1\right]$ with stop-grad operator $\mathbf{sg}$. The terms $\bz^{\phi}_{t_{t-s}}$ and $f_\theta\left(\bz_{t_i}, t_i\right)$ are defined as follow: 
\begin{align}
    &f_\theta\left( \bz_{t_i}, t_i \right) = \bz_{t_i} - t_i * v_\theta\left( \bz_{t_i}, t_i \right) , \\
    &\bz^{\phi}_{t_{i-s}} = \bz_{t_i} - (t_i-t_{i-s}) * v_\phi\left( \bz_{t_i}, t_i \right).
\end{align}
Solely applying consistency distillation \cite{song2023consistency} on latent space presents two challenges: (1) one-step synthesis produces blurry images, which would significantly degrade the FID metric - this observation aligns with findings in \cite{luo2023latent}; (2) when sampling with few-step, the student model generates oversaturated images, as illustrated in \Cref{fig:loss_ablate}. These limitations could due to the statistical difference between latent and pixel space. To address these limitations, we propose to use GAN and Reflow techniques.

\minisection{Blurry outputs of one-step generation.} We realize that one-step images produced by student model is blurry as seen in first row of \Cref{fig:loss_ablate}. Since single-step image still contain coarse structure information, we propose to apply GAN to further boost the sharpness of one-step images. To ensure that the one-step images already contain coarse structure, we start applying GAN loss after several iterations of consistency distillation. This is similar to the warm-up technique of VQGAN \cite{esser2021taming} to reduce the training instability. The proposed GAN is as follow:
\begin{equation}
    \mathcal{L}_{GAN} = \mathcal{D}_{adv}(f_\theta\left( \bz_{1}, 1 \right), z_{0}), \label{eq:L_GAN}
\end{equation}
where $f_\theta\left( \bz_{1}, 1 \right)$ is the student's one-step generated image.

\begin{figure}[t]
    \centering
    \includegraphics[width=\linewidth]{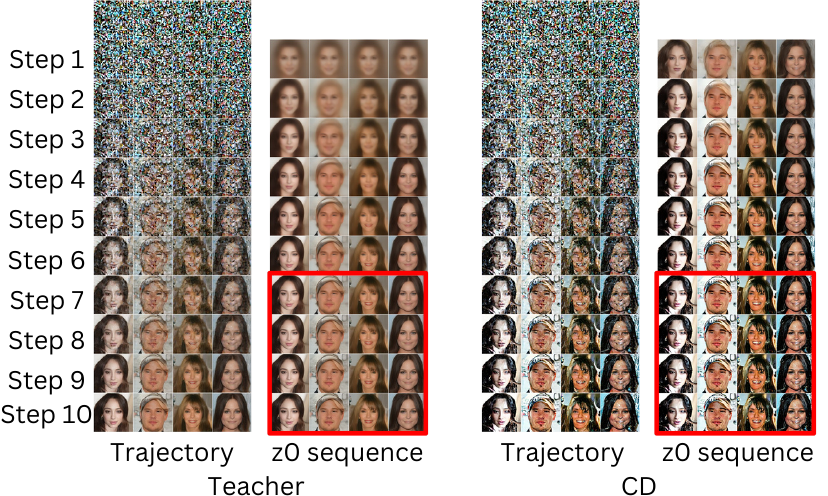}
    \caption{Trajectory of 10 NFEs Euler sampling of vanilla flow matching (teacher model) and CD model.}
    \label{fig:motivation}
\end{figure}

\minisection{Oversaturated outputs of few-step generation.} As shown in \Cref{fig:motivation}, we realize that $f_\theta\left( \bz_{t_i}, t_i \right)$ become over-saturated as $t_i \leq 0.4$. Besides, we observe that when $t_i$ is small, $f_\theta(x_{t_i}, {t_i})$ can be well approximated by $z_0$, which is not hold for large $t_i$. Therefore, we can use diffusion loss $\Vert f_\theta(x_{t_i}, t_i)-x_0 \Vert_2^2$ instead of $\Sigma_{t_i<0.4}\Vert f_{\tilde{\theta}}(x_{t_{i-s}}, t_{i-s})-f_\theta(x_{t_i}, t_i)\Vert_2^2$. Furthermore, the reason for not using consistency loss for small $t_i$ is that the value of $f_{\tilde{\theta}}(x_{t_{i-s}}, t_{i-s})-f_\theta(x_{t_i}, {t_i}) \approx 0$, therefore the update gradient for it is minimal leading to subliminal gradient update on small $t_i$. By using truncated consistency loss (\cref{eq:L_CD}), we observe less oversaturated synthesis for a few-step generation.

\begin{equation}
    \resizebox{\linewidth}{!}{%
    $\mathcal{L}_{CD} = 
    \begin{cases}
        \mathbf{E}_{t_i, \bz_{t_i}} \left[\norm{f_{\tilde\theta} \left( \bz^{\phi}_{t_{i-s}}, t_{i-s} \right) - f_\theta\left(\bz_{t_i}, t_i\right) }^2_2 \right] \text{if $t_i > 0.4$}\\
        \mathbf{E}_{t_i, \bz_{t_i}} \left[\norm{\bz_0 - f_\theta\left(\bz_{t_i}, t_i\right) }^2_2 \right] \text{if $t_i \leq 0.4$}
    \end{cases}$
    }
    \label{eq:L_CD}
\end{equation}

However, truncated $\mathcal{L}_{CD}$ cannot fully eliminate oversaturared limitation as shown in \Cref{fig:loss_ablate}. In addition, since the proposed GAN loss only enhance the quality of one-step image, the inconsistency still exists between one-step and few-step images as shown in second row of \Cref{fig:loss_ablate}. 

To better reduce saturated effect and improve the consistency between one-step and few-step synthesized images, we propose a reflow loss motivated from RectifiedFlow \cite{liu2022rectified, liu2023instaflow}. This loss directly self-corrects the flow estimation, given a reliable estimate of one-step source $f_\theta(z_1, 1)$ as below:
\begin{equation}
    \mathcal{L}_{RF} = \mathbf{E}_{t_i, \hat\bz_{t_i} } \left[\norm{\mathbf{sg}(f_\theta\left( \bz_{1}, 1 \right)) - f_\theta\left(\hat\bz_{t_i}, t_i\right) }^2_2 \right], \label{eq:L_RF}
\end{equation}
with $\hat\bz_{t_i} = (1-t_i) * \mathbf{sg}(f_\theta\left( \bz_{1}, 1 \right)) + t_i*\bz_1$.

\begin{table}[t]
\centering
  \small

  \begin{tabular}{lccccc}
    \toprule
     Method & NFE$\downarrow$ &  FID30K$\downarrow$ & CLIP$\uparrow$ & P$\uparrow$ & R$\downarrow$ \\
     \toprule
    2-RF (teacher) & 25 & 11.08 & - & - & - \\
    \toprule
    Guided Distill & 1 & 37.30	& 0.270 &	- &	- \\
    UFOGen	& 1 & 12.78	& - &	- &	- \\
    SD Turbo & 1& 	16.10 &	 \textbf{0.330} &	\textbf{0.65} &	0.35 \\
    \toprule
    LCM	& 1 & 35.56 &	0.24 & 	-	& - \\
    InstaFlow-0.9B & 1 &	13.10 &	0.280 &	0.53 &	0.45 \\
    InstaFlow-1.7B & 1 &	11.83 &	- &	- &	- \\
    2-RF & 1 & 36.83 & 0.282 & 0.30 & 0.33  \\
    Ours-0.9B & 1 &	\textbf{11.91} &	\underline{0.312}	& \underline{0.54} &	\textbf{0.47} \\
    Ours-0.9B & 2 &	\textbf{11.46} &	\underline{0.315}	& \underline{0.57} &	\textbf{0.46} \\
    \bottomrule
    \end{tabular}
\caption{Text-to-image results on zero-shot COCO2014.}
\label{tab:t2i}
\end{table}

Rectified Flow \cite{liu2022rectified} requires to sample $\bz_0$ by using multi-step generation before applying rectified flow technique. This process costs both time and memory for generating high-quality images from teacher model. Unlike RectifiedFlow, we directly use one-step image $f_\theta\left( \bz_{1}, 1 \right)$ for rectified flow technique instead of multi-step image. Interestingly, one-step images are not oversaturated and are high quality due to GAN loss as seen in second row of \Cref{fig:loss_ablate}. Consequently, our proposed reflow loss effectively addresses the oversaturation issue in few-step sampling while enhancing consistency between few-step and one-step generation through the straightness penalty of the rectified flow loss, as illustrated in the third row of \Cref{fig:loss_ablate}.

\minisection{Bi-directional Consistency Distillation.} In consistency distillation, the $L_{CD}$ objective forces the output $f_\theta(z_{t_i}, t_i)$ to close to $f_{\theta}(z_0, 0) \approx x_0$, the high-quality source at the end of the denoising process (due to $||f_\theta\left( \bz_{t_i}, t_i \right) - f_\theta\left( \bz_{0}, 0 \right)||_2^2 \leq \sum_{i}||f_\theta\left( \bz_{t_i}, t_i \right) - f_{\theta}\left( \bz_{t_{i-1}}, t_{i-1} \right)||_2^2 = L_{CD}$). However, thanks to the GAN objective (\cref{eq:L_GAN}), we can generate high-quality one-step samples $f_{\theta}(z_1, 1)$, at the start of the denoising process.  Therefore, incorporating the bidirectional loss ensures that $f_{\theta}(z_{t_i}, t_i)$ receives quality signals from both endpoints of the denoising process, thus enhancing the consistency at both directions. The bi-directional objective is written below:
\begin{equation}
    \mathcal{L}_{BI} = \mathbf{E}_{t_i, \bz_{t_i}} \left[\norm{f_{\tilde\theta} \left( \bz^{\phi}_{t_{i+s}}, t_{i+s} \right) - f_\theta\left(\bz_{t_i}, t_i\right) }^2_2 \right]. \label{eq:L_Bi}
\end{equation}

Importantly, it is activated only when high-quality 1-NFE outputs is ensured by GAN loss, allowing beneficial signals to guide student training and avoiding poor-quality information in early training stages. 

With the proposed loss terms, our distill student could be able to generate high quality images in both one and few step setting, refer to \Cref{fig:nfe_samples}. Our overall distillation framework is briefly described by the \Cref{alg:utrain}.

\section{Experiment}
\begin{table}[t]
\centering
\small
\begin{tabular}{lcccc}
  \toprule
   Method & NFE$\downarrow$ &  FID5K$\downarrow$ & CLIP$\uparrow$ & Time (s) \\
   \toprule
  2-RF (teacher) &	25 &	21.50 &	0.315 & 0.88 \\
  \toprule
  Guided Distill.	&1	&37.20	&0.275	&0.09 \\
  2-RF	&1	&47.00	&0.271	&0.09 \\
  UFOGen	&1	&22.50	&0.311	&0.09 \\
  InstaFlow-0.9B	&1	&23.40	&0.304	&0.09 \\
  InstaFlow-1.7B	&1	&22.40	&0.309	&0.12 \\
  Ours-0.9B	&1	&\textbf{22.09}	&\textbf{0.313}	&0.09 \\
  \midrule
  Guided Distill.	&2	&26.00	&0.297	&0.13 \\
  2-RF	&2	&31.30	&0.296	&0.13 \\
  Ours-0.9B	&2	&\textbf{21.20} &\textbf{0.317} &0.13 \\	
  \bottomrule
\end{tabular}
\caption{Text-to-image results on zero-shot COCO2017.}
\label{tab:t2i_coco2017_5k}
\end{table}

\subsection{Self-Corrected Flow Distillation For Unconditional Generation} \label{ssec:distill}

\minisection{Training details.} Our experiments are conducted on CelebA-HQ 256 for pretrained latent flow matching model from LFM \cite{dao2023flow}. We modify and use the discriminator architecture from \cite{phung2023wavelet}. Our distillation procedure uses 200 training epochs with learning rate 1e-5 for both discriminator and student. The default ema rate $\mu$ is 0.9, $t_{trunc}$ is 0.4 and $t_{skip}$ is 0.1. The loss weight $(\lambda_{GAN, RF, BI})$ and warm-up iteration $(N_{GAN, RF, BI})$ are set to $(0.1, 0.1, 0.1)$ and $(0, 1000, 1000)$. For sampling process, we use for Euler solver by default.

\begin{table}[t]
\centering
\footnotesize
\begin{tabular}{l c c}
\toprule
Model & NFE$\downarrow$ & FID$\downarrow$  \\
\toprule
\multicolumn{3}{c}{One-Step} \\ 
\midrule
LFM  & 1 & 200.13 \\
LFM+ Rectified  & 1 & 18.03 \\
LFM+ Rectified + Distill  & 1 & 12.95 \\
LFM+ CD & 1 & 41.34 \\
\midrule
\rowcolor{pink!60} Ours & 1 & 8.06 \\
\midrule
\multicolumn{3}{c}{Multi-Step} \\ 
\midrule
LFM+ Rectified + Distill  & 2 & 30.85  \\
LFM+ CD & 2 & 23.56 \\
\midrule
\rowcolor{pink!60} Ours  & 2 & 7.67 \\
\bottomrule
\end{tabular}
\caption{Quantitative results on CelebA-HQ 256.}
\label{tab:distill}
\end{table}

\begin{figure}[h]
\centering
\includegraphics[width=0.9\linewidth]{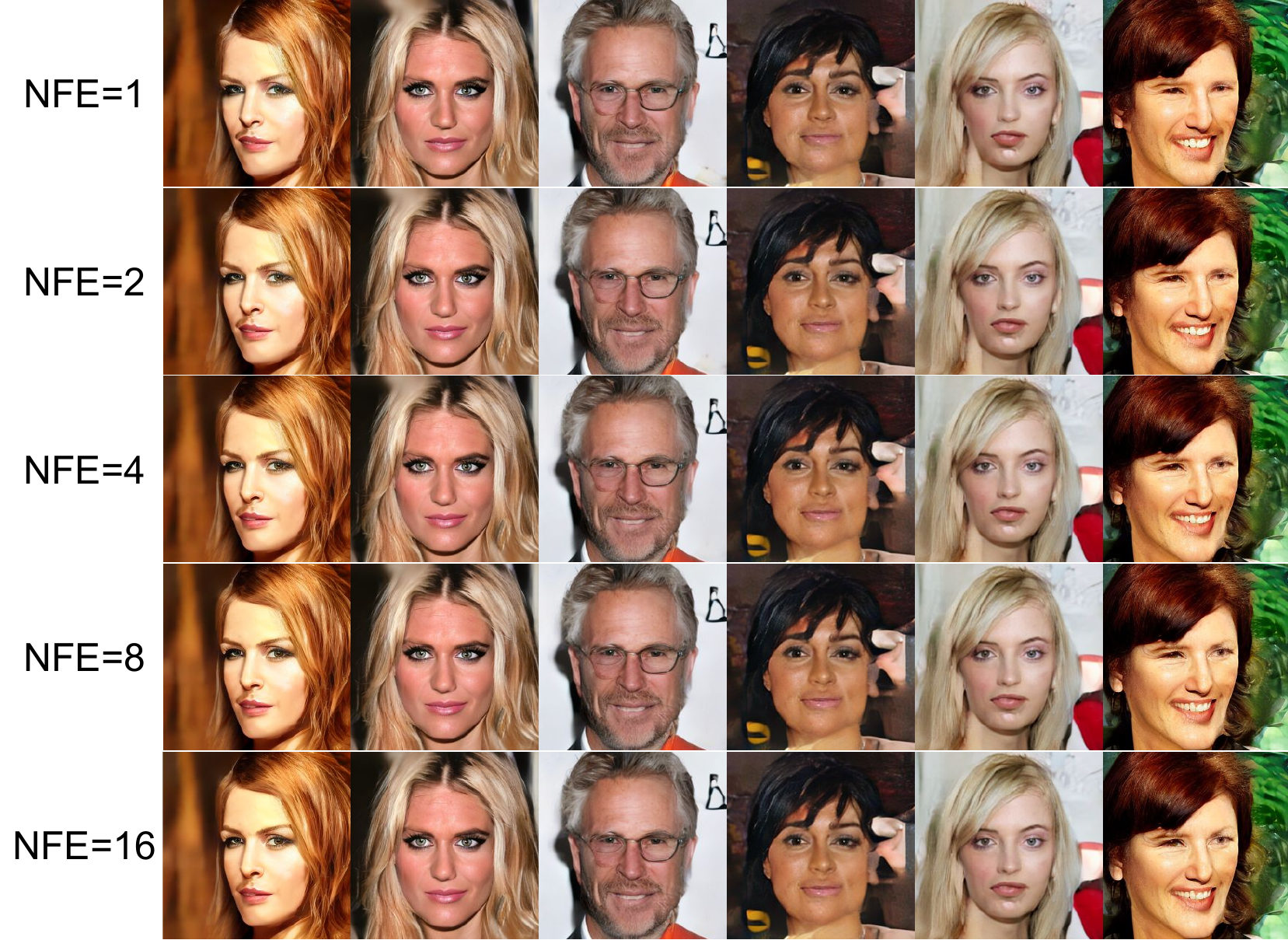}
\caption{Varying NFEs on CelebA-HQ. Increasing NFEs accentuates details and sharpness in generated faces without oversaturation issues.}
\label{fig:nfe_samples}
\end{figure}

\minisection{Experimental results.} We compare our method with 2 baselines Rectified Flow \cite{liu2022rectified, liu2023instaflow} and Consistency Distillation \cite{song2023consistency, luo2023latent}. The reason for choosing these baseline is that the technique allows both one and few-step sampling scheme. For rectified flow, we follow rectified distillation framework \cite{liu2022rectified} which comprises of three stage: data generation, rectified flow and distillation. We firstly create a set of 50,000 pairs ($\bz_1$, $\bz_0$) using 500 steps Euler solver, where $\bz_1$, $\bz_0$ are random noise and synthesized image correspondingly. We then train rectified flow for 50 epochs on the synthesized set. Finally, we perform distillation stage by directly mapping from $\bz_1$ to $\bz_0$ in 10 epochs. For Consistency Distillation, we follow \cite{song2023consistency} implementation and distill model for 50 epochs on CelebA-HQ 256 real dataset.

\begin{figure}[t]
    \centering
    \includegraphics[width=\linewidth]{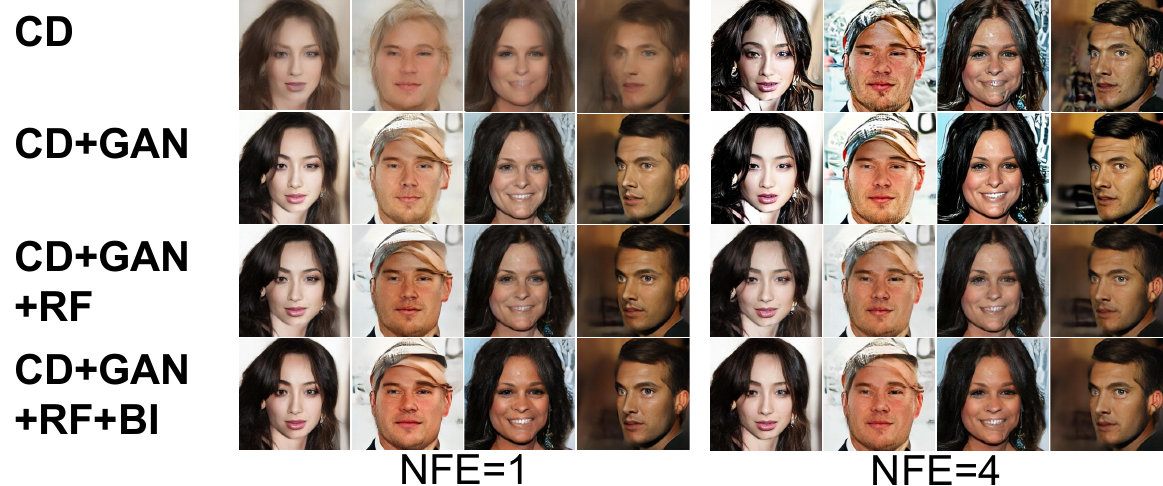}
    \caption{Qualitative of loss choice: 1 NFE vs 4 NFE}
    \label{fig:loss_ablate}
\end{figure}

The experiment result is reported in Table \ref{tab:distill}. For one-step generation, our approach achieves 8.06 FID which outperforms all the baselines. For two-step sampling, the same observation is also hold. Our method's FID is 7.67 compared to Rectified Flow 30.85 and Consistency Distillation 23.56. Notably, 2-step FID of Rectified Distillation is higher than 1-step counterpart because the third stage mapping from $\bz_1$ to $\bz_0$ hurts the multistep sampling ability. 
Refer to \Cref{fig:teaser} for quality comparison between Rectified Distillation framework and our proposed framework. For the quality comparison with Consistency Distillation, please check the first and last row of \Cref{fig:loss_ablate}. These results underscore the efficacy of self-corrected flow distillation which not only produce high quality one-step generation but also maintaining high-quality with multiple-step sampling. Furthermore, \Cref{fig:nfe_samples} and \Cref{fig:teaser} demonstrate the consistency generation of our framework with both few-step and one-step sampling produces same image given same noise input.



\subsection{Self-Corrected Flow Distillation For Text-to-Image Generation} \label{ssec:distill_t2i}

\minisection{Evaluation metrics.} We evaluate our text-to-image model using a \textbf{``zero-shot"} framework, wherein the model is trained on one dataset and tested on another, ensuring a robust assessment of generalization capabilities. Our evaluation encompasses three critical dimensions: image quality, diversity, and textual fidelity.
The primary metric for assessing image quality is the Fréchet Inception Distance (FID) \cite{heusel2017gans}. In addition to FID, we employ precision and recall \cite{kynkaanniemi2019improved} as a complementary metric to assess image quality and diversity. For measuring the alignment between generated images and their corresponding text prompts, we use the CLIP score \cite{radford2021learning}. Following \cite{liu2023instaflow, gu2023boot,sauer2023adversarial,luo2023latent,sauer2023stylegan,kang2023gigagan}, we employ the MS COCO-2014 validation set and MSCOCO-2017 as our standard zero-shot text-to-image benchmarks. For MSCOCO-2014, we generate samples from the first 30,000 prompts, while for MSCOCO-2017, we use the first 5,000 prompts. 

\minisection{Training details.} For our experiments, we employ a two-stage rectified flow (2-RF) model as the teacher, ensuring a fair comparison with InstaFlow. Our training process utilizes 2 million samples from the LAION dataset with an aesthetic score larger than 6.25. The architecture of our discriminator is based on a UNet-encoder design, augmented with an additional head, drawing inspiration from the UFOGen model. The model undergoes training for 18,000 iterations, with a consistent learning rate of 1e-5 applied to both the generator and discriminator components. To maintain consistency, all other hyperparameters are aligned with the configuration used in our CelebAHQ experiments.

\minisection{Experimental results.} Table \ref{tab:t2i} presents our zero-shot text-to-image generation results on COCO2014, comparing our method with state-of-the-art approaches like 2-RF \cite{liu2023instaflow}, Guided Distillation \cite{meng2023distillation}, UFOGen \cite{xu2024ufogen}, SD Turbo \cite{sauer2023adversarial}, LCM \cite{luo2023latent}, and InstaFlow \cite{liu2023instaflow}.

Model Ours-0.9B achieves the best FID score of 11.91 with just one step, surpassing all other methods, including the larger InstaFlow-1.7B. This demonstrates our approach's efficiency in generating high-quality images with minimal computation. For text-image alignment, our model's CLIP score of 0.312 is second only to SD Turbo (0.330), indicating high relevance to input prompts. It also shows a good balance between precision (0.54) and recall (0.47) for one-step generation, suggesting diverse yet accurate image generation. Increasing to two steps further improves performance, with an FID of 11.46 and a CLIP score of 0.315. This showcases our approach's scalability and ability to leverage additional computational steps for enhanced quality.

\begin{table}[t]
    \footnotesize
    \centering
        \begin{tabular}{l | c c c c c c}
        \toprule
        Loss &NFE=$1$ &NFE=$2$ &NFE=$4$ &NFE=$8$ &NFE=$16$\\
        \toprule
        $\cL_{CD}$ &41.34 &23.56 &38.75 &51.34 &53.69 \\
        $+ \cL_{GAN}$ &11.55 &19.82 &38.29 &49.45 &54.56\\
        $+ \cL_{RF}$ &10.62 &11.07 &11.8 &12.68 &12.66\\
        \rowcolor{pink!60} $+\cL_{BI}$ &8.06 &7.67 &7.68 &7.63 &7.24\\
        \bottomrule
        \end{tabular}
    \caption{Ablation of our Self-Corrected Flow Distillation. FID is used for all experiments (Lower is better).}
    \label{tab:ab_distill_loss}
\end{table}

On the other hand, table \ref{tab:t2i_coco2017_5k} shows the results of our zero-shot evaluation on the COCO2017 dataset. Here, we observe similar trends to the COCO2014 results, with our model outperforming other methods in both one-step and two-step generation scenarios. For one-step generation, our model achieves the best FID score of 22.09 and the highest CLIP score of 0.313 among all compared methods. This performance is particularly impressive considering that our model matches or exceeds the quality of models with more parameters (e.g., InstaFlow-1.7B). When increasing to two steps, our model further improves its performance, achieving an FID of 21.20 and a CLIP score of 0.317. This not only outperforms other two-step methods but also surpasses the quality of models using many more steps, such as 2-RF with NFE=25. Importantly, our model maintains competitive inference times, with 0.09 seconds for one-step and 0.13 seconds for two-step generation, which is comparable to other efficient methods and significantly faster than multi-step approaches. For qualitative result, please refer to \Cref{fig:t2i_teaser}.

\begin{table}[t]
    \small
    \centering
    \setlength{\tabcolsep}{1mm} 
    \begin{tabular}{l c c c c c}
        \toprule
        $(\lambda_{GAN,RF,BI})$ &NFE=$1$ &NFE=$2$ &NFE=$4$ &NFE=$8$ &NFE=$16$ \\
        \toprule
        $(1.0, 1.0, 1.0)$ &30.26 &29.71 &29.22 &30.12 &35.57 \\
        $(0.4, 0.4, 0.4)$ & 12.22 & 12.07 & 11.66 & 11.06 & 9.95 \\
        $(0.2, 0.2, 0.2)$ &9.83 &9.65 &9.43 &9.33 &8.82 \\
        \rowcolor{pink!60} $(0.1, 0.1, 0.1)$ &8.06 &7.67 &7.68 &7.63 &7.24 \\
        $(1.0, 0.1, 0.1)$ &11.2 &11.38 &11.26 &11.25 &12.55 \\
        $(0.1, 1.0, 1.0)$ &35.22 &31.68 &30.8 &31.44 &33.12 \\
        \bottomrule
        \end{tabular}
        \caption{Ablation of weighting loss terms. By default, we use our best hyper-params like $t_{trunct} = 0.4$, $\mu=0.9$, and $t_{skip}=0.1$ in case one of those is not explicitly mentioned in the table.}
        \label{tab:ab_distill_weighting}
\end{table}

\subsection{Ablation Studies For Self-Corrected Flow Distillation}

We conduct extensive ablation studies on our distilled model, with results presented in \Cref{tab:ab_distill_hyper} and \Cref{tab:ab_distill_weighting}. Specifically, Table \ref{tab:ab_distill_hyper} demonstrates that model performance is mostly influenced by three key parameters: the time-truncated threshold $t_{trunct}$ in $\mathcal{L}_{CD}$, the EMA decay $\mu$, and the time-skip threshold $t_{skip}$. Our findings indicate that optimal results are achieved when $t_{trunct}$ is within the range $[0.2, 0.5)$ and $t_{skip}$ is approximately 0.1. Notably, an EMA decay of $\mu = 0.9$ consistently yields superior performance, particularly for many-step generation.

Table \ref{tab:ab_distill_weighting} explores the impact of weights on the GAN loss, reflow loss, and bidirectional consistency loss. Our results indicate that lower weights (around 0.1 to 0.2) for these components lead to optimal performance, highlighting the critical role of precise loss balancing in our framework.

Qualitative results across various NFEs are presented in \Cref{fig:nfe_samples}. Furthermore, \Cref{fig:loss_ablate} illustrates the progressive improvements achieved by each component of our method:
\begin{itemize}
    \item Consistency distillation alone (row 1) can lead to increased contrast and statistical shift at higher NFEs, explaining the higher FID scores observed in \Cref{tab:ab_distill_loss}.
    \item The addition of GAN and TCD losses (row 2) address the blurriness in one-step generation but does not fully resolve oversaturation in multistep outputs.
    \item Reflow loss (row 3) enforces consistency between one-step and many-step generations, mitigating the oversaturation issue.
    \item The bidirectional term (row 4) further enhances the consistency generation across one and few-step sampling. 
\end{itemize}

These observations underscore the statistical discrepancies between pixel and latent spaces, which manifest as blurriness in one-step generation and oversaturation in few-step generation. Our proposed method effectively mitigates these issues, as evidenced by \Cref{fig:nfe_samples}, where high-quality images are produced across various NFEs without oversaturation.


\begin{table}[t]
    \centering
    \setlength{\tabcolsep}{1mm} 
    \begin{tabular}{l c c c c c}
        \toprule
         &NFE=$1$ &NFE=$2$ &NFE=$4$ &NFE=$8$ &NFE=$16$ \\
        \toprule
        \multicolumn{6}{c}{Time-truncated threshold in $\cL_{CD}$} \\
        \toprule
        $t_{trunc} = 0.$ &8.50 &8.89 &9.42 &9.86 &10.97 \\
        $t_{trunc} = 0.1$ &8.44 &8.80 &9.36 &9.87 &11.03 \\
        $t_{trunc} = 0.2$ &7.50 &7.56 &7.75 &7.75 &8.22 \\
        \rowcolor{pink!60} $t_{trunc} = 0.4$ &8.06 &7.67 &7.68 &7.63 &7.24 \\
        $t_{trunc} = 0.5$ &9.55 &9.26 &9.24 &9.12 &8.28 \\
        \toprule
        \multicolumn{6}{c}{EMA decay} \\
        \toprule
        $\mu = 0.9$ &8.06 &7.67 &7.68 &7.63 &7.24 \\
        $\mu = 0.95$ &7.79 &7.90 &8.07 &8.00 &7.93 \\

        \toprule
        \multicolumn{6}{c}{Time-skip threshold} \\
        \toprule
        
        \rowcolor{pink!60} $t_{skip} = 0.1$ &8.06 &7.67 &7.68 &7.63 &7.24 \\
        $t_{skip} = 0.2$ &11.86 &11.53 &11.42 &11.27 &10.97 \\
        
        \bottomrule
        \end{tabular}
        \caption{Ablation of time-truncated threshold, EMA decay, and time-skip threshold. By default, we use our best hyper-params like $t_{trunc} = 0.4$, $\mu=0.9$, and $t_{skip}=0.1$ in case one of those is not explicitly mentioned in the table.}
        \label{tab:ab_distill_hyper}
        
\end{table}

\section{Conclusion}
This work presents Self-Corrected Flow Distillation ensuring consistent, high-quality generation in both one-step and few-step sampling. Our method successfully mitigates the limitation of latent consistency distillation, including blurry single-step and oversaturated multi-step samples. Our extensive experiments on CelebA-HQ and text-to-image generation tasks demonstrate substantial improvements over existing methods, achieving superior FID and visual quality for both one and few-steps sampling. 

\section{Acknowledgements}
Research partially funded by research grants to Prof. Dimitris Metaxas from NSF: 2310966, 2235405, 2212301, 2003874, 1951890, AFOSR 23RT0630, and NIH 2R01HL127661.

\bibliography{aaai25}

\newpage
\appendix

\section{Pseudo code of time-skip generation}

In \Cref{alg:timeskip}, we show the pseudo-code of time-skip threshold $t_{skip}$ used in $\mathcal{L}_{BI}$ and $\mathcal{L}_{CD}$.

\begin{algorithm}[h]
\caption{Time-skip generation}
\label{alg:timeskip}
\DontPrintSemicolon
\KwData{current time $t_i \in [0, 1]$, time-skip threshold $t_{skip} \in [0, 1]$}
\Comment*[l]{skip range}
$r_{s} \gets clip(t_i, 0, t_{skip})$ \;
$r_{k} \gets clip(1.0 - t_i, 0, t_{skip})$ \;
\Comment*[l]{random skip step}
$\delta_{s} \gets rand(0, 1) * r_{s}$ \;
$\delta_{k} \gets rand(0, 1) * r_{k}$ \;
$t_{i-s} \gets t_i - \delta_{s}$ \Comment*[l]{use in $\cL_{CD}$}
$t_{i+k} \gets t_i + \delta_{k}$ \Comment*[l]{use in $\cL_{BI}$}

\KwResult{$t_{i-s}, t_{i+k}$}
\end{algorithm}

\begin{figure}[h]
    \centering
    \includegraphics[width=0.98\linewidth]{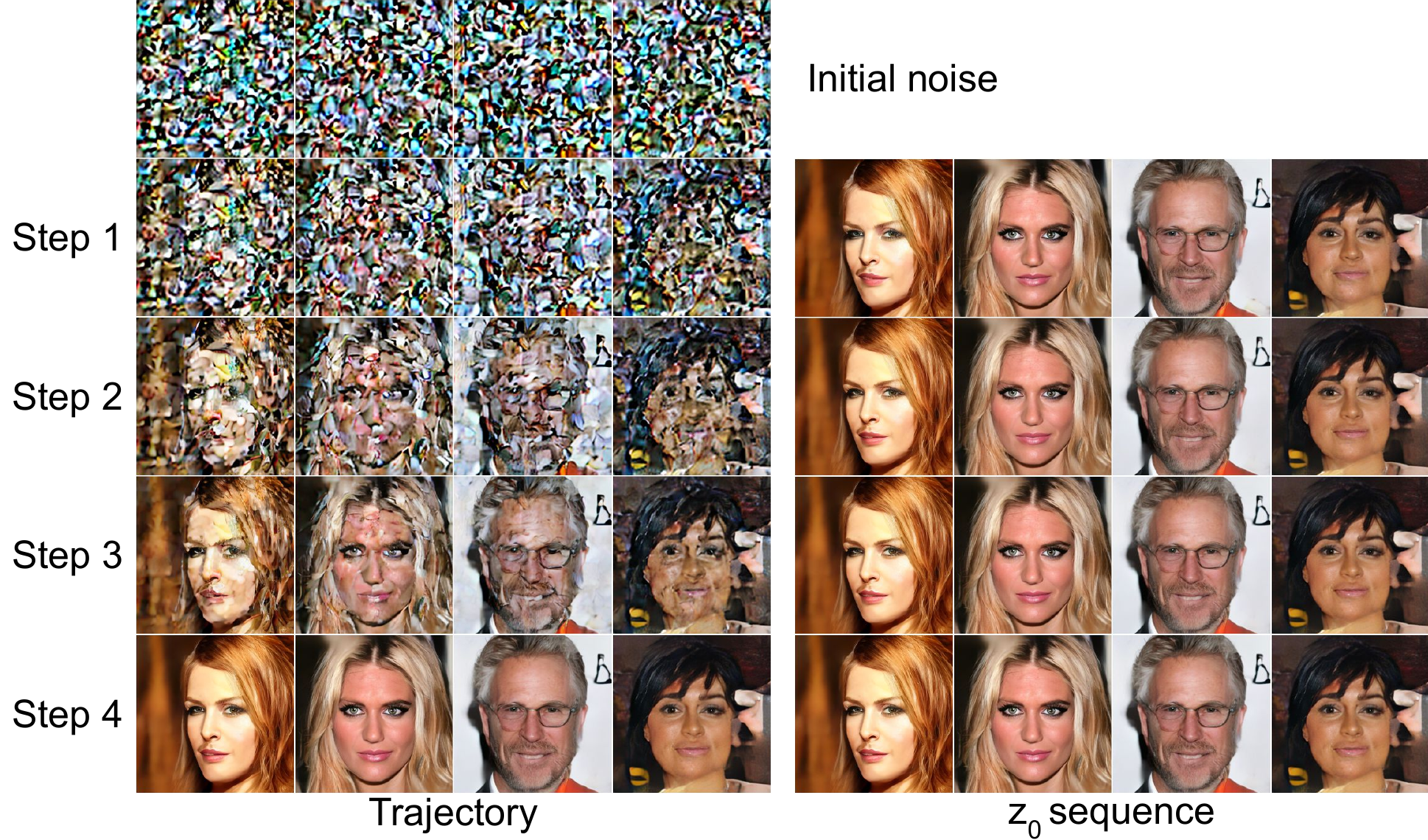}
    \vspace{-1mm}
    \caption{Trajectories of four-step sampling. $z_0$ sequence means that at each step $t$, the approximated clean output $\hat{z_0}$ is directly estimated by one-step Euler update.}
    \label{fig:traj}
\end{figure}

\section{Qualitative results}
We show the trajectory and $z_0$-prediction in \Cref{fig:traj}. As seen in the figure, the generation results at each step remain mostly identical, highlighting the effectiveness of our distillation method in NFE-consistent generation.

\begin{figure*}
    \centering
    \includegraphics[width=1\linewidth]{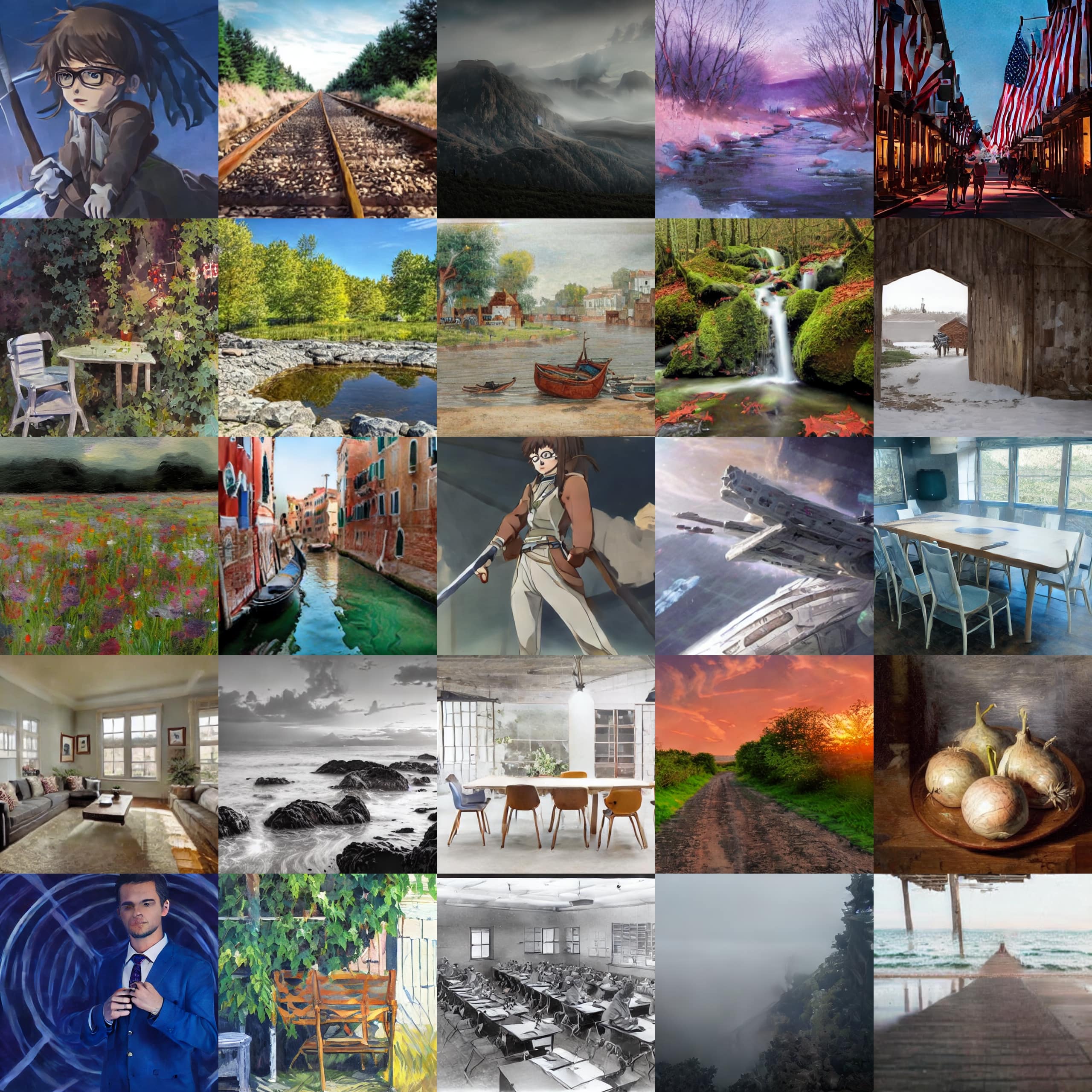}
    \caption{Uncurated samples of our text-to-image model using NFE=1}
    \label{fig:1nfe_v1}
\end{figure*}

\begin{figure*}
    \centering
    \includegraphics[width=1\linewidth]{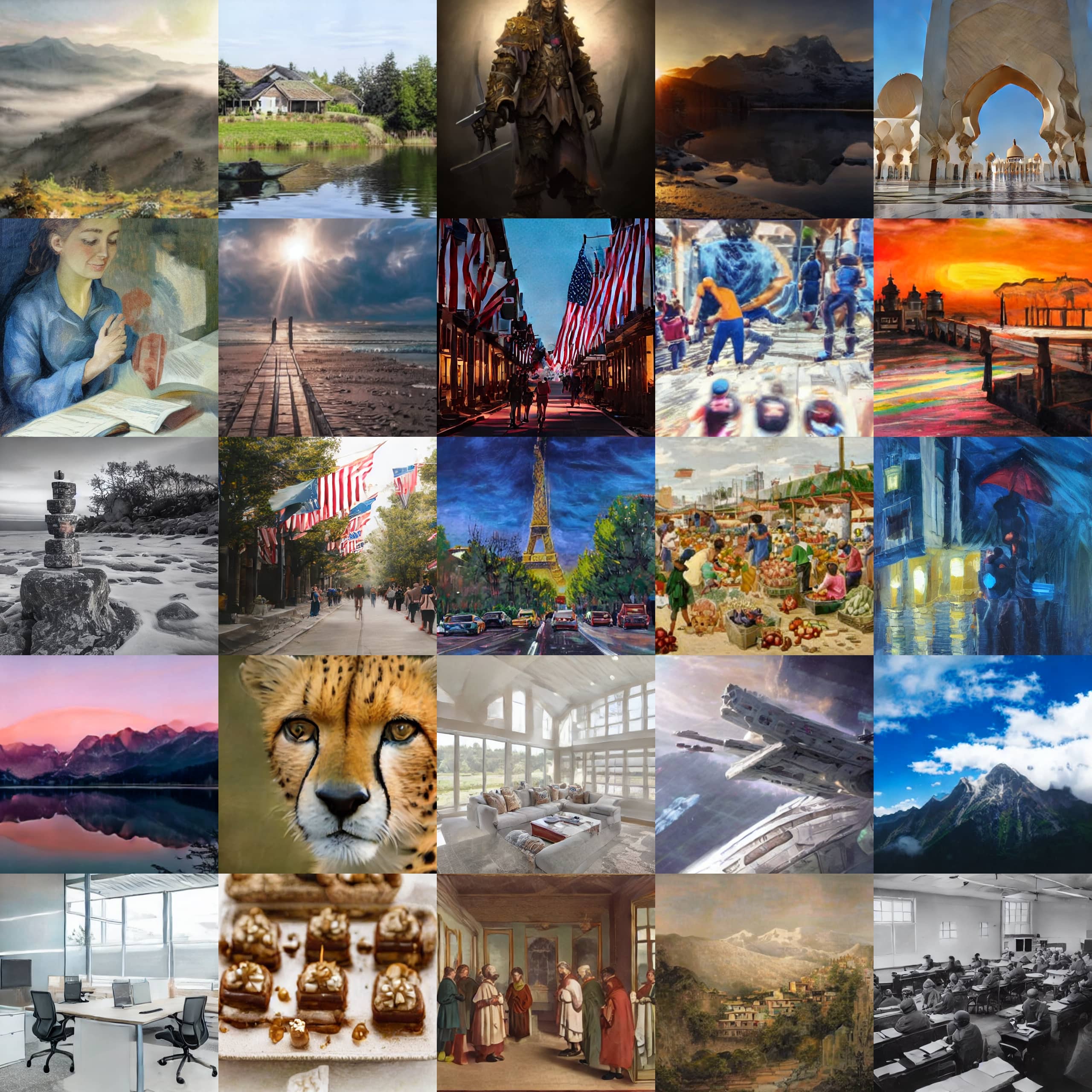}
    \caption{Uncurated samples of our text-to-image model using NFE=1}
    \label{fig:1nfe_v2}
\end{figure*}

\begin{figure*}
    \centering
    \includegraphics[width=1\linewidth]{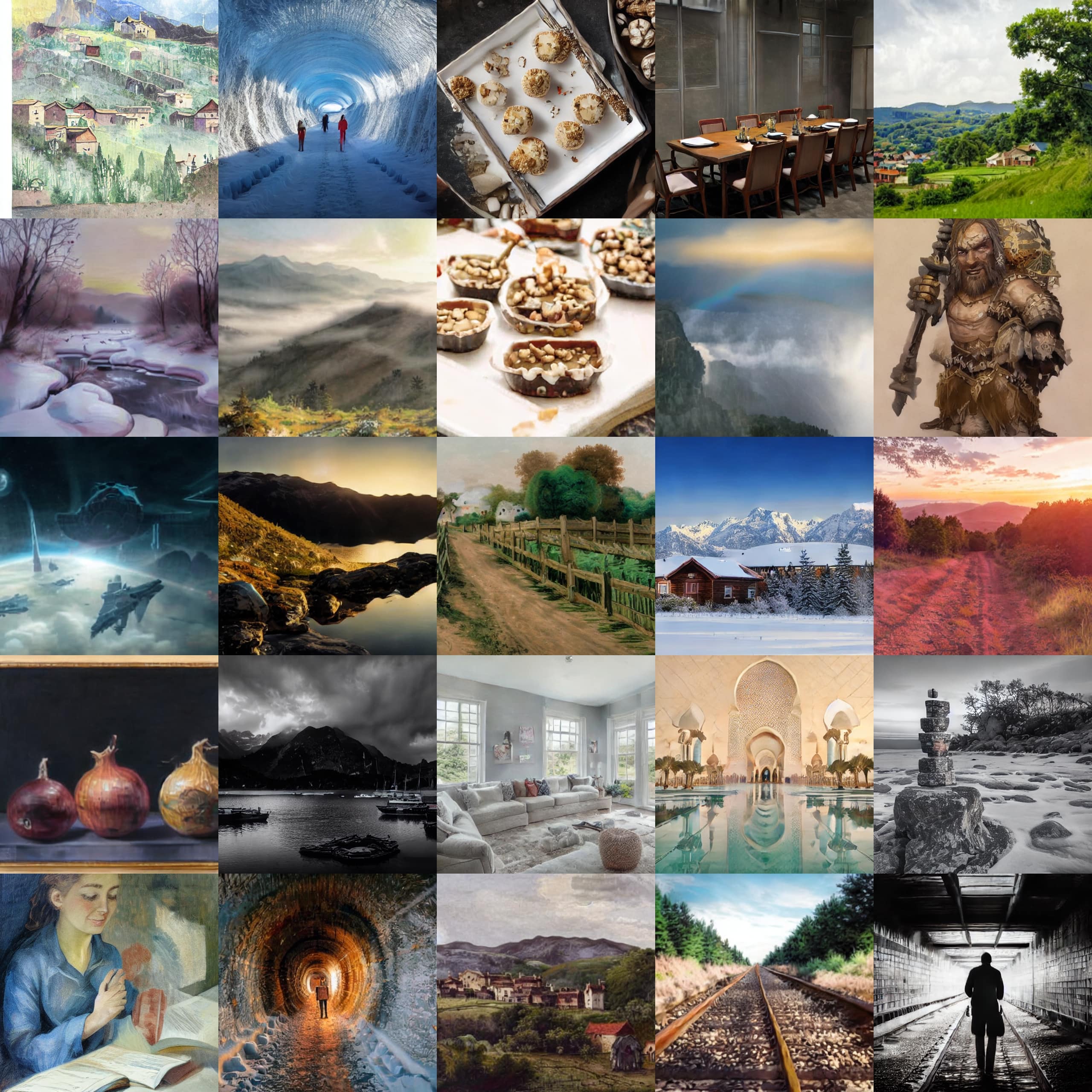}
    \caption{Uncurated samples of our text-to-image model using NFE=1}
    \label{fig:1nfe_v3}
\end{figure*}

\begin{figure*}
    \centering
    \includegraphics[width=1\linewidth]{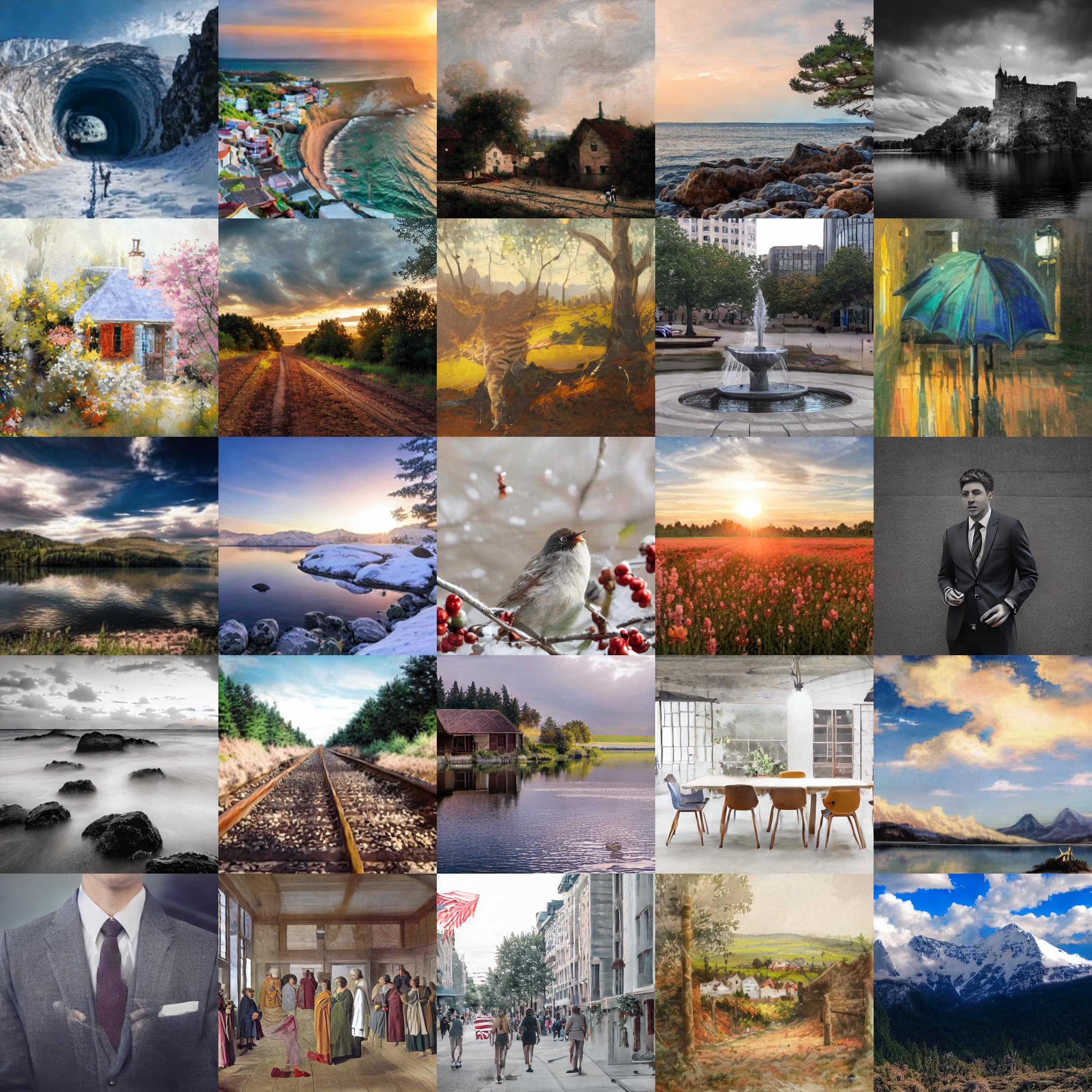}
    \caption{Uncurated samples of our text-to-image model using NFE=1}
    \label{fig:1nfe_v4}
\end{figure*}

\begin{figure*}
    \centering
    \includegraphics[width=1\linewidth]{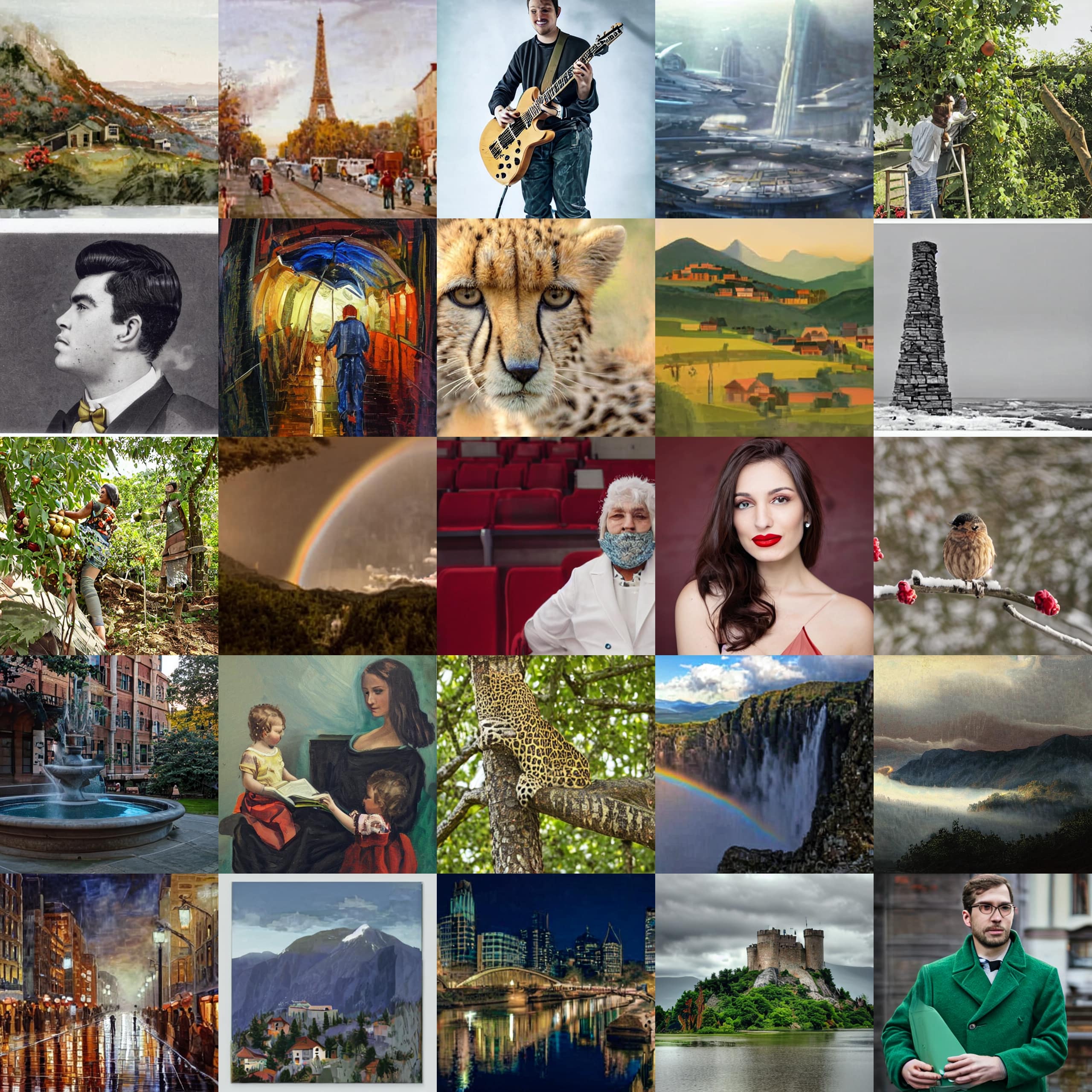}
    \caption{Uncurated samples of our text-to-image model using NFE=2}
    \label{fig:2nfe_v1}
\end{figure*}

\begin{figure*}
    \centering
    \includegraphics[width=1\linewidth]{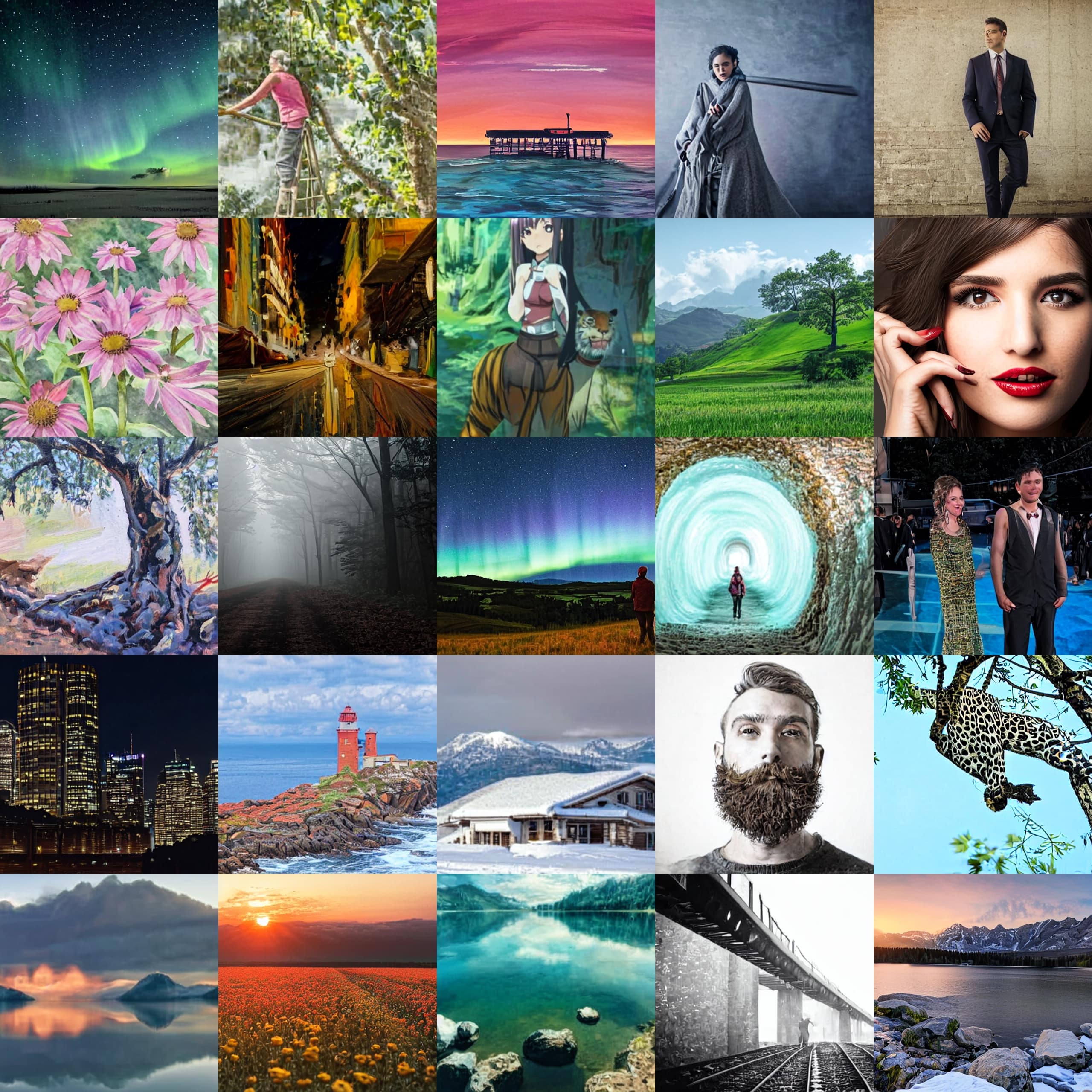}
    \caption{Uncurated samples of our text-to-image model using NFE=2}
    \label{fig:2nfe_v2}
\end{figure*}

\begin{figure*}
    \centering
    \includegraphics[width=1\linewidth]{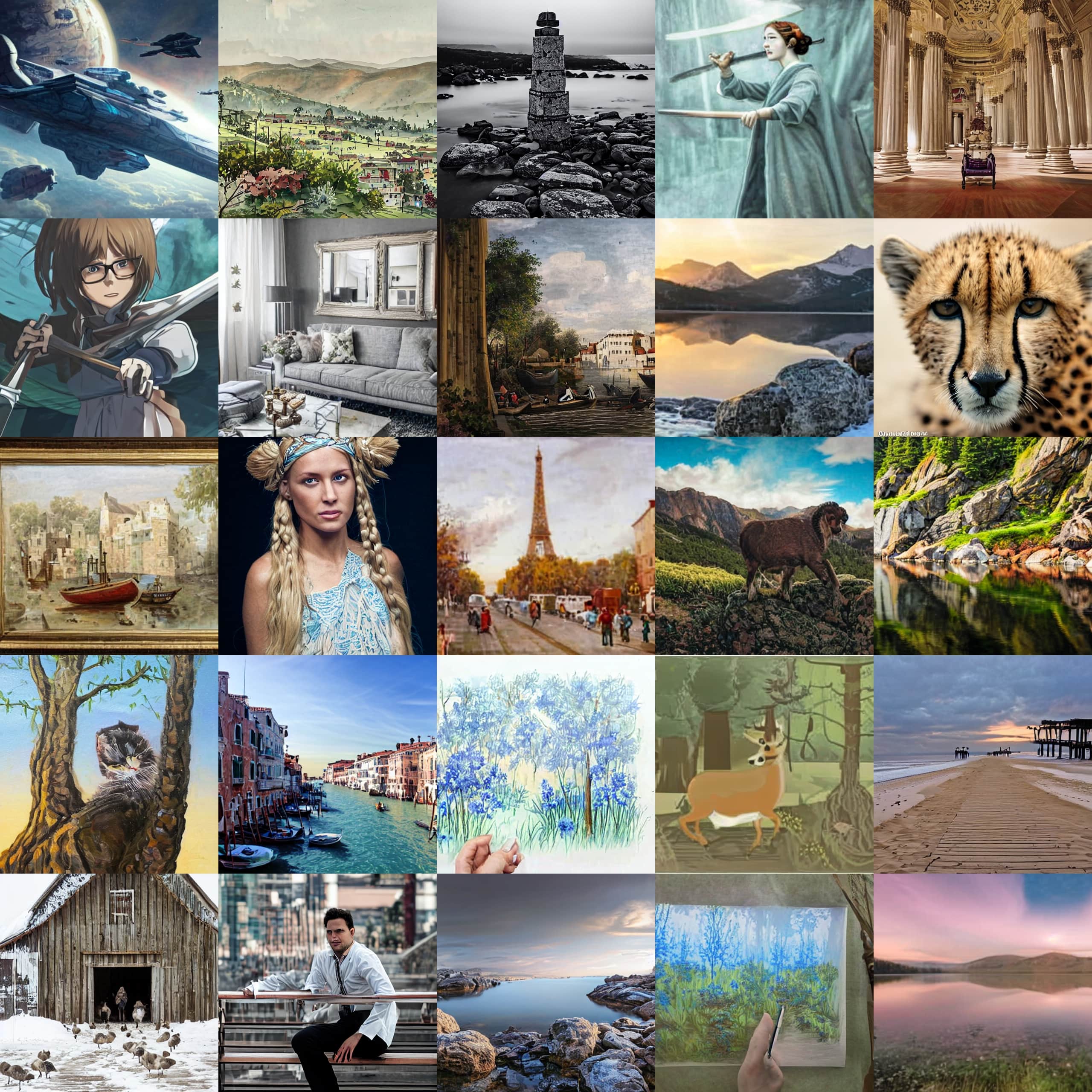}
    \caption{Uncurated samples of our text-to-image model using NFE=2}
    \label{fig:2nfe_v3}
\end{figure*}

\begin{figure*}
    \centering
    \includegraphics[width=1\linewidth]{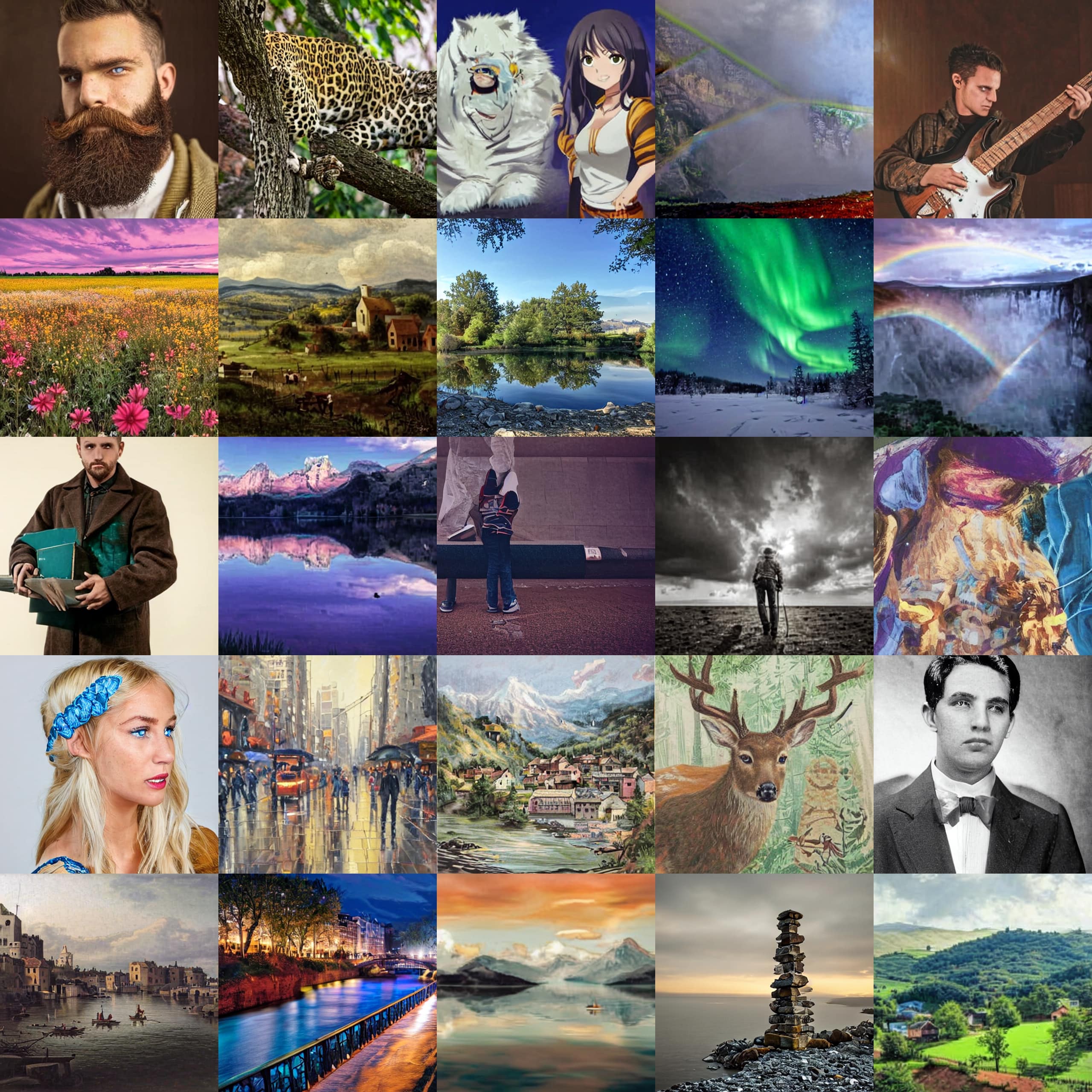}
    \caption{Uncurated samples of our text-to-image model using NFE=2}
    \label{fig:2nfe_v4}
\end{figure*}

We present comprehensive visual qualitatives of our distilled text-to-image diffusion model's capabilities across different sampling configurations. Figures \ref{fig:1nfe_v1} through \ref{fig:1nfe_v4} showcase a diverse array of images generated using only one denoising step (NFE=1), demonstrating the model's efficiency in producing high-quality results with minimal computational overhead. Figures \ref{fig:2nfe_v1} through \ref{fig:2nfe_v4} display outputs produced with two denoising step (NFE=2).

\end{document}